%% file: main.tex
\documentclass[10pt,twocolumn,letterpaper]{article}

\usepackage[pagenumbers]{cvpr} 

\input{preamble}
\definecolor{cvprblue}{rgb}{0.21,0.49,0.74}

\definecolor{rank_red}{HTML}{FFB3B3}
\definecolor{rank_orange}{HTML}{FFD9B3}
\definecolor{rank_yellow}{HTML}{FFFFB3}
\newcommand{\best}[1]{{\cellcolor{rank_red} #1}}
\newcommand{\sbest}[1]{{\cellcolor{rank_orange} #1}}
\newcommand{\tbest}[1]{{\cellcolor{rank_yellow} #1}}

\usepackage[pagebackref,breaklinks,colorlinks,allcolors=cvprblue]{hyperref}

\def\confName{CVPR}
\def\confYear{2026}

\title{Multi-view Consistent 3D Gaussian Head Avatars \\ `without'  Multi-view Generation}
\vspace{-0.2in}
\usepackage{mathrsfs}
\usepackage{algorithm}
\usepackage{algorithmic}
\usepackage{amsmath}
\usepackage{amssymb}
\usepackage{multirow}
\usepackage{capt-of,etoolbox}
\usepackage[most]{tcolorbox}
\usepackage{booktabs}
\usepackage{multirow}
\usepackage{subcaption}
\usepackage{colortbl}
\usepackage{xcolor} 
\newtcolorbox{insightbox}[1]{%
  colback=white,         
  colframe=gray!20,     
  colbacktitle=cvprblue, 
  coltitle=white,        
  title=Insight~#1,      
  arc=2mm,               
  boxrule=0.8pt,
  left=2mm,right=2mm,top=1mm,bottom=1mm,
}

\author{%
Aviral Chharia,\ \ Fernando De la Torre\\
Carnegie Mellon University \\
{\tt \{achharia, ftorre\}@cs.cmu.edu}\\
}
\begin{document}
\twocolumn[{
\renewcommand\twocolumn[1][]{#1}
\maketitle
\vspace{-3em}

\begin{center}
    \captionsetup{type=figure}
    \centering
    \vspace{-0.5em}
    \includegraphics[width=1.03\textwidth]{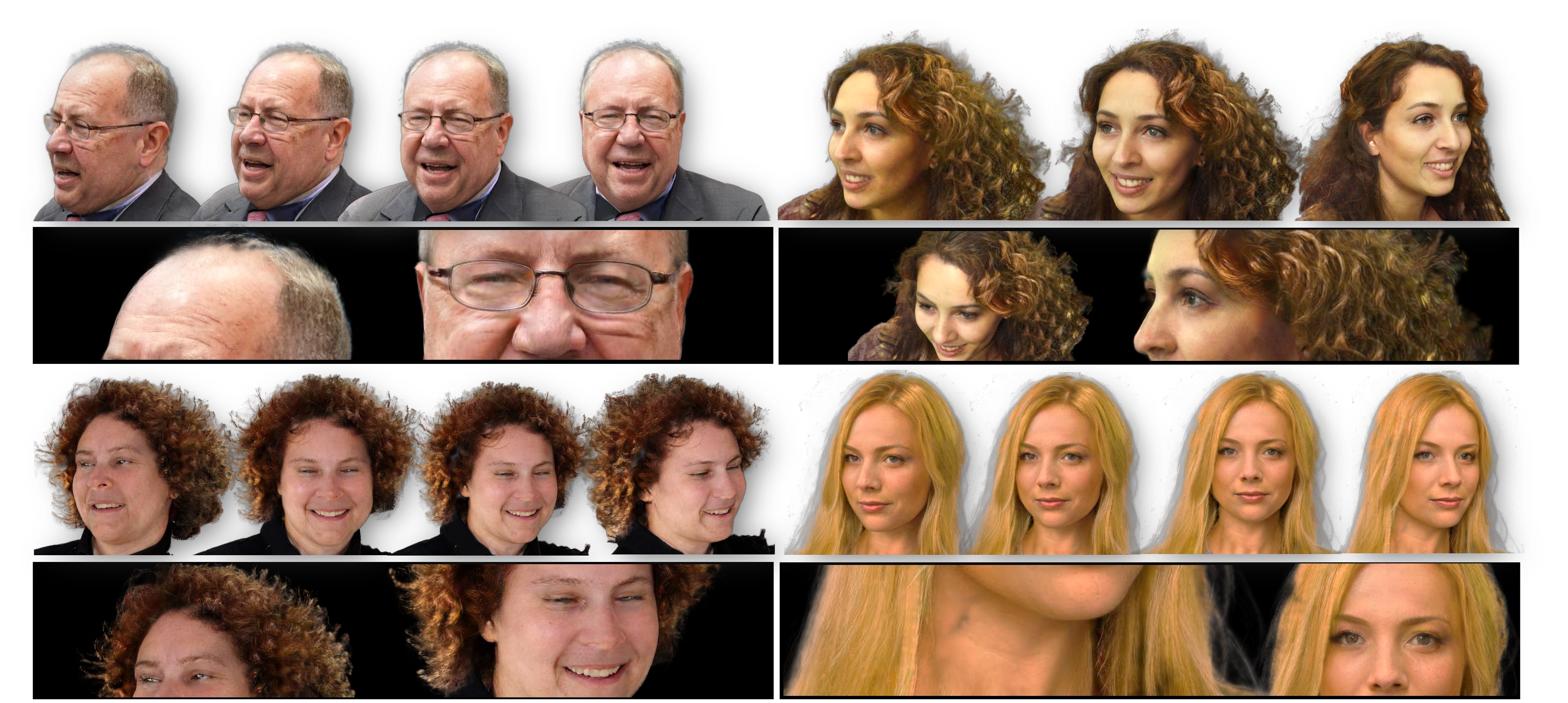}
    \vspace{-1.75em}
    \caption{\textit{\textbf{MVCHead}} achieves state-of-the-art for unconditional generation of high fidelity, multi-view consistent 3D Gaussian head avatars in ``minimal resource setting'', without requiring intermediate views, or even 3D data. The generated Gaussian heads capture complex textures and fine facial micro-structure, including wrinkles, hair wisps, ear rims, lip contours, skin blemishes, eyes, and accessories.}
    \vspace{0.5em}
     \label{fig:1_teasor}
\end{center}}]
\input{sec/0_abstract}
\input{sec/1_intro}
{
    \small
    \bibliographystyle{ieeenat_fullname}
    \bibliography{main}
}
\end{document}

%% file: sec/0_abstract.tex
\begin{abstract}
High-fidelity 3D Gaussian head avatar generation is critical for applications such as AR/VR, telepresence, and digital humans. Existing methods depend on multi-view datasets, 3D captures, or intermediate 2D view synthesis. In contrast, we learn both conditional and unconditional 3D head models from randomly sampled 2D images alone, without using multi-view data, 3D supervision, or intermediate view generation. We introduce MVCHead, a single-shot state space model that enforces multi-view consistency (MVC) directly in the 3D representation while regressing 3D Gaussians under these constraints. At its core, we propose a Hierarchical State Space (HiSS) block that progressively refines Gaussians from coarse to fine, while capturing long-range dependencies. Within each HiSS block, we modify Mamba's standard unidirectional scan with the proposed Hierarchical Bi-directional State Scan (HiBiSS) that aligns recurrence with the axes along which multi-view inconsistencies are strongest. Finally, we design an SE(3) Multi-view Critic that judges whether a set of self-renders arises from a single underlying 3D configuration, rewarding cross-view pixel alignment without observing real multi-view pairs. MVCHead achieves state-of-the-art perceptual quality, surpasses prior methods in both texture and geometric consistency, and maintains comparable shape consistency. To demonstrate scalability, we release FaceGS-10K, the first large-scale dataset of ready-to-use 3D Gaussian head assets for training and evaluation of 3D head models. Project Page and code: \url{https://humansensinglab.github.io/MVCHead/}
\end{abstract}

%% file: sec/1_intro.tex
\vspace{-1em}
\section{Introduction}
\label{sec:intro}
High-fidelity 3D Gaussian head avatars have become central to AR/VR, telepresence, digital characters, and large-scale content creation in film and games~\cite{zheng2025headgap, yan2024gaussian, zhang2025hravatar, martinez2024codec, huang2026blurry, aneja2024gaussianspeech, kirschstein2025avat3r, liu2024human}. These applications demand vast numbers of realistic yet non-identifiable 3D head avatars that are consistent across views but correspond to no real individual--avoiding privacy concerns and enabling rapid content creation. Generating such assets in a minimal-resource setting (e.g., from 2D images alone) is practically important, especially for studios that cannot afford dense multi-view capture rigs or high-end 3D scanning. Moreover, multi-view diffusion pipelines that first synthesize intermediate views are computationally heavy and often require additional training data. Motivated by these constraints, we explore this `minimal-resource setting'.

\begin{figure}[t]
    \centering
    \includegraphics[width=0.5\textwidth]{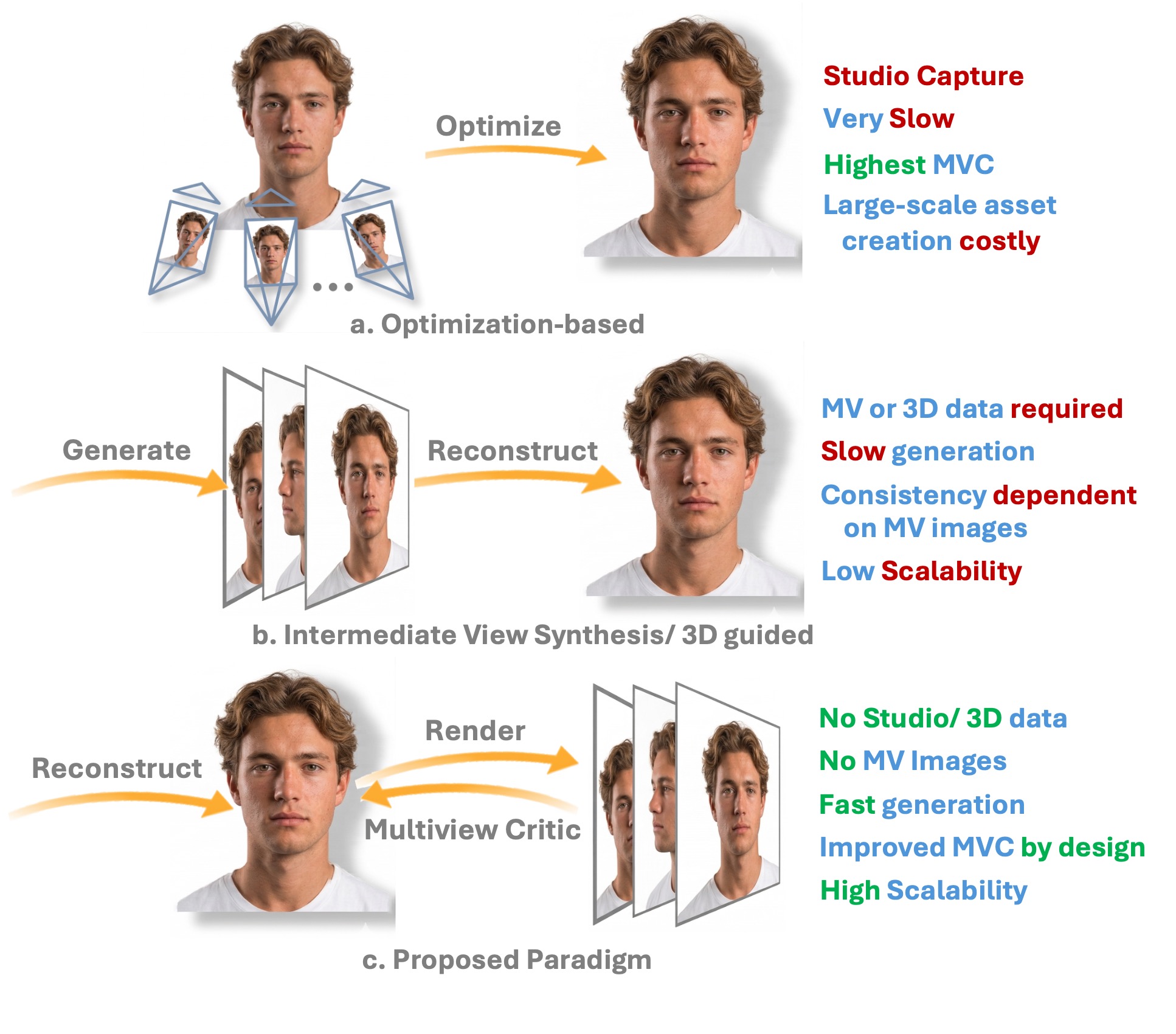}
    \vspace{-2.5em}
    \caption{\textbf{Motivation.} Paradigms for 3D Gaussian head avatar generation. (a) Requires expensive studio captures; (b) Synthesizes intermediate views before reconstruction; (c) Learns an unconditional 3D Gaussian head directly from 2D images w/o intermediate generation or even 3D data.}
    \vspace{-1.5em}
    \label{fig:2_motivation}
\end{figure}

Recent work on 3D Gaussian head avatar generation falls into three broad categories that differ primarily in supervision, data requirements, and scalability (see Fig.~\ref{fig:2_motivation}). \textit{First}, multi-view optimization-based methods~\cite{aneja2025scaffoldavatar, qian2024gaussianavatars, giebenhain2024npga, teotia2024gaussianheads, wang20253d, chen2024mixedgaussianavatar, dhamo2024headgas} reconstruct a full 3D head from high-resolution studio-captured sequences with dense multi-view coverage. These pipelines, using datasets such as NeRSemble~\cite{kirschstein2023nersemble} or RenderMe-360~\cite{pan2023renderme} (with $\sim10^4$ frames per subject), achieve impressive photorealism and strong MVC (see Fig.~\ref{fig:2_motivation}(a)). However, reliance on costly capture setups and heavy per-subject optimization limits scalability.

A \textit{second} class of methods~\cite{lyu2025facelift,taubner2025cap4d,zhou2025zero,deng2024portrait4d, deng2024portrait4d_v2, yin2025facecraft4d, galanakis2025spinmeround, gu2024diffportrait3d, gu2025diffportrait360, liao2025soap} encompasses multi-view diffusion approaches that start from a single image and first synthesize intermediate views, typically including side views of the subject via off-the-shelf image or video diffusion models (see Fig.~\ref{fig:2_motivation}(b)). A separate reconstructor then lifts these images into a 3DGS representation~\cite{kerbl20233d}. While fidelity is high, MVC becomes tightly coupled to intermediate view quality: pixel-aligned cross-view losses are not optimized since there is no end-to-end differentiability, and identity drift persists: tiny per-view deviations (e.g., subtle shifts in hair, ear contours, or jawline shading) may not correspond to any consistent 3D explanation. Moreover, generating dense intermediate views per asset is computationally prohibitive at scale.

A \textit{third} line of works~\cite{kirschstein2024gghead, hyun2024gsgan, barthel2025cgs, barthel2024gaussian} includes feed-forward 3D generators that directly produce 3D Gaussian head avatars in an end-to-end differentiable manner. These methods aim for unconditional generation of 3D Gaussian heads from a learned prior, enabling the creation of diverse, non-existent identities while avoiding per-subject optimization. GGHead~\cite{kirschstein2024gghead}, GS-GAN~\cite{hyun2024gsgan}, and CGS-GAN~\cite{barthel2025cgs} improve stability, yet enforcing MVC without explicit multi-view supervision remains open, particularly in minimal-resource settings when the model never observes real multi-view pairs. In this work, we tackle this highly challenging minimal-resource setting (see Fig.~\ref{fig:2_motivation}(c)): achieving large-scale, real-time synthesis of multi-view consistent 3D Gaussian head avatars via a single-shot, end-to-end differentiable model that operates \textit{(i) without generating intermediate views and (ii) without relying on 3D ground truth}.

To address this, we introduce \textit{MVCHead}, a novel state space model tailored to this setting. To the best of our knowledge, MVCHead is the \textit{first} to leverage state space modeling for 3D Gaussian head generation. It takes a latent code and produces a complete set of 3D Gaussians in a single forward pass. MVCHead consists of a series of Hierarchical State Space (HiSS) blocks that organize Gaussians in a hierarchy and guide finer levels through offsets anchored to coarser parent Gaussians. Within each HiSS block, we apply the proposed Hierarchical Bi-directional State Scan (HiBiSS), which enforces grid-aligned coherence to reconcile typical view-to-view drift. Finally, we propose an SE(3) Multi-view Critic that rewards cross-view pixel alignment, inducing multi-view consistency by design. Taken together, MVCHead combines architectural improvements with a learned consistency critic to generate 3D Gaussian head avatars of high visual quality and strong multi-view consistency (see Fig.~\ref{fig:1_teasor}). Our main contributions include:
\begin{itemize}
    \item We highlight the challenge of MVC and analyze how it can be induced by design, arguing that intermediate view generation is counterproductive for scalability. We propose an SE(3) Multi-view Critic that rewards cross-view pixel alignment without real multi-view pairs.

    \item We introduce \textit{MVCHead}, the first to leverage visual Mamba for 3D Gaussian head generation: a fast, single-shot state space model that directly predicts Gaussians and improves MVC in unconditional 3D head synthesis.
    
    \item We modify Mamba's traditional unidirectional scan into a Hierarchical Bi-directional State Scan (HiBiSS) that aligns recurrence with principal axes of multi-view drift.

    \item MVCHead surpasses the state-of-the-art in perceptual quality and along all three MVC axes, achieving superior texture and geometric consistency while maintaining comparable shape consistency.

    \item We release FaceGS-10K, a large-scale dataset of ready-to-use 3D Gaussian heads for large-scale training, benchmarking, and evaluation of 3D-aware head models.
\end{itemize}

\section{Related Works}

\subsection{3D Gaussian Head Avatars}

\noindent \textbf{Multi-view optimization-based methods.} A large body of work~\cite{aneja2025scaffoldavatar, qian2024gaussianavatars, giebenhain2024npga, teotia2024gaussianheads, wang20253d, chen2024mixedgaussianavatar, dhamo2024headgas} reconstructs detailed 3D heads by optimizing Gaussians against dense, high-resolution studio-captured multi-view video sequences such as RenderMe-360~\cite{pan2023renderme} and NeRSemble~\cite{kirschstein2023nersemble}, which largely guarantee MVC. GaussianAvatars~\cite{qian2024gaussianavatars} rigs Gaussians to FLAME~\cite{li2017learning}; SplattingAvatar~\cite{shao2024splattingavatar} leverages monocular video; GaussianHeadAvatars~\cite{xu2024gaussian} and MonoGaussianAvatar~\cite{chen2024monogaussianavatar} exploit multi-view data but from relatively sparse or monocular views. These set an upper bound on quality but offer low scalability due to expensive capture and per-subject optimization.\vspace{0.25em}

\noindent \textbf{Multi-view diffusion methods.} These models~\cite{lyu2025facelift,zhou2025zero,deng2024portrait4d, deng2024portrait4d_v2, yin2025facecraft4d, galanakis2025spinmeround, taubner2025cap4d, taubner2025mvp4d} generate 3D head avatars from a single input image by first synthesizing several intermediate views~\cite{gu2024diffportrait3d, gu2025diffportrait360, liao2025soap} via off-the-shelf image or video diffusion~\cite{lyu2025facelift} and subsequently reconstructing the avatar. Zero-1-to-A~\cite{zhou2025zero}, FaceLift~\cite{lyu2025facelift}, Cap4D~\cite{taubner2025cap4d}, FaceCraft4D~\cite{yin2025facecraft4d}, SpinMeRound~\cite{galanakis2025spinmeround}, and Portrait4D~\cite{deng2024portrait4d, deng2024portrait4d_v2} have achieved impressive fidelity in this two-stage setup. Cap4D~\cite{taubner2025cap4d} and FaceCraft4D~\cite{yin2025facecraft4d} target 4D controllability; Portrait4D~\cite{deng2024portrait4d, deng2024portrait4d_v2} variants improve identity stability across expression and view changes; FaceLift~\cite{lyu2025facelift} couples multi-view diffusion with Gaussian reconstruction. While fidelity is high, MVC hinges on the intermediate view generator; pixel-aligned cross-view losses are not optimized end-to-end, and identity drift across synthesized views persists. Moreover, dense multi-view generation for each asset is computationally prohibitive at scale. Other works leverage monocular or multi-view videos~\cite{tang2025gaf, giebenhain2024mononphm, kirschstein2024diffusionavatars, zhang2025fate, feng2025gpavatar, li2025rgbavatar}.\vspace{0.25em}

\noindent \textbf{Feed-forward and other methods.} These methods~\cite{he2025lam,chu2024generalizable,li2025panolam, oroz2025perchead, chu2024gpavatar,kirschstein2024gghead, hyun2024gsgan, barthel2025cgs} generate avatars directly in 3D through a feed-forward mapping from latent codes to Gaussians. Recent large Gaussian reconstruction models such as LAM~\cite{he2025lam}, GAGAvatar~\cite{chu2024generalizable}, PanoLAM~\cite{li2025panolam}, PercHead~\cite{oroz2025perchead}, and GPAvatar~\cite{chu2024gpavatar} reintroduce end-to-end differentiability but rely on large-scale video datasets~\cite{xie2022vfhq}, multi-view data from Cafca~\cite{buehler2024cafca}, or studio-collected 3D data~\cite{kirschstein2023nersemble, pan2023renderme} to impose MVC. GGHead~\cite{kirschstein2024gghead} 
uses a 2D CNN model to predict Gaussian attributes in a UV-template head and regularizes geometry via a total-variation loss. Hyun~\textit{et al.}~\cite{hyun2024gsgan} introduce hierarchical Gaussians to stabilize training; Barthel~\textit{et al.}~\cite{barthel2025cgs} address the challenge of view conditioning. Despite this progress, enforcing MVC without paired multi-view supervision remains a key bottleneck.

\subsection{State Space Models}
State Space Models (SSMs) originate from classical linear dynamical systems and Kalman filtering~\cite{kalman1960new}. Gu~\textit{et al.} introduced the modern Structured State Space Sequence (S4) family, which demonstrated strong long-range dependency modeling~\cite{gu2021efficiently, gu2021combining}. Mamba~\cite{gu2024mamba} extends S4 by replacing its fixed hidden-space projection matrices with an input-dependent selective projection mechanism. Recent variants~\cite{liu2024vmamba, zhu2024vision, li2024mamba} adapt SSM scanning to 2D and higher-dimensional inputs. Hybrid Mamba-Transformer architectures~\cite{hatamizadeh2025mambavision} have achieved SOTA performance on ImageNet-1K~\cite{deng2009imagenet} classification and multiple vision tasks~\cite{chharia2025mv, dong2024hamba}. Despite this, the use of SSMs in 3D generative modeling remains largely unexplored. Gamba~\cite{shen2025gamba} combines Mamba with 3DGS for single-view reconstruction but shows limited texture quality; MVGamba~\cite{yi2024mvgamba} targets simple objects for content creation rather than human heads. MVCHead is the first to leverage SSMs for 3D head avatar generation. We use SSMs to align recurrence with the axes along which multi-view inconsistencies manifest, making state space propagation instrumental in improving MVC.

\section{MVCHead}
We aim to learn a generative mapping from a latent code $z$ to a 3D head, represented as a set of anisotropic Gaussians~\cite{kerbl20233d}. Unlike methods that rely on expensive studio captures or additional view synthesis, we operate in a minimal-resource setting, supervising solely on 2D images.\vspace{0.25em}

\noindent\textbf{Notation and Preliminaries.} For a latent code $z\sim \mathcal{N}(0, I)$, we generate a set of anisotropic Gaussians~\cite{kerbl20233d}, $\mathcal{S}_\theta(z)=\{g_i\}_{i=1}^{N}$. Each individual Gaussian $g_i$ is defined by the tuple $g_i = (\mu_i, s_i, q_i, \alpha_i, c_i)$. Here, $\mu_i \in \mathbb{R}^3$ denotes the 3D center, $s_i \in \mathbb{R}^3_+$ encodes positive axis-aligned scales, $q_i \in \mathbb{H}$ is a unit quaternion defining a rotation matrix $R(q_i) \in SO(3)$, $\alpha_i \in (0,1)$ is an opacity value, and $c_i \in [0,1]^3$ is an RGB color. We fix the Gaussian budget at N$=240\text{K}$, which is sufficient for high-fidelity modeling of facial features. A differentiable splatting renderer $\mathcal{R}$ maps $\mathcal{S}_\theta(z)$ and a camera pose $T\in SE(3)$ to an image: $\mathbf{I} = \mathcal{R}(\mathcal{S}_\theta(z), T) \in \mathbb{R}^{H \times W \times 3}$. Crucially, the only supervision comes from 2D images sampled from large face corpora; these images provide a texture and appearance distribution but no ground truth cross-view correspondences.\vspace{0.25em}

\begin{figure*}[t]
    \centering
    \includegraphics[width=\textwidth]{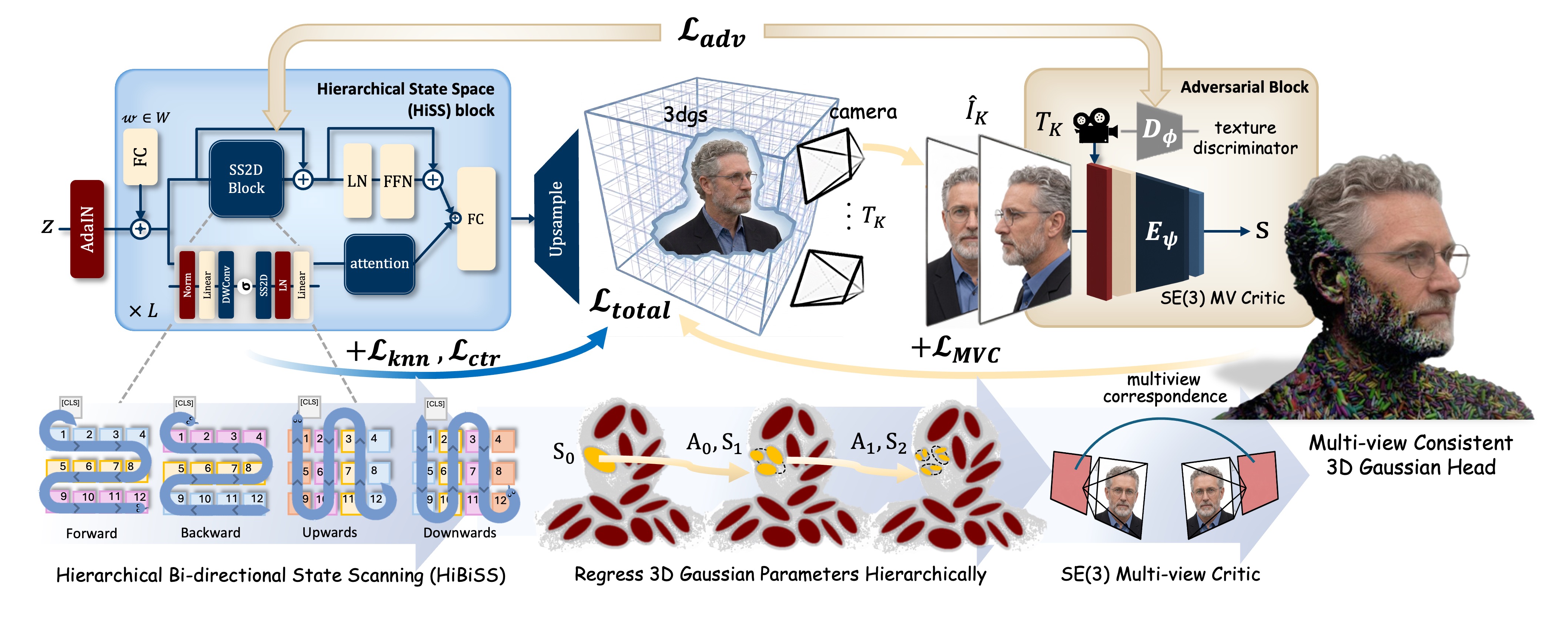}
    \vspace{-2.5em}
    \caption{\textbf{Model Architecture.} MVCHead along with its key proposed components, including HiSS blocks which hierarchically regress the 3D Gaussian parameters (Gaussian $S_0$ becomes the anchor $A_0$ for computing the next Gaussian $S_1$, and so on), and perform Hierarchical Bi-directional State Scan (HiBiSS) in all directions, and the SE(3) Multi-view Critic, which enforces MVC.}
    \vspace{-1em}
    \label{fig:3_model_architecture}
\end{figure*}

\noindent \textbf{Overview.} The proposed model architecture is illustrated in Fig.~\ref{fig:3_model_architecture}. We build on the transformer-based GSGAN~\cite{hyun2024gsgan}, making three key departures: a novel Dual-Mixer architecture leveraging the state space blocks; the proposed HiBiSS scan; and the SE(3) Multi-view Critic as an explicit MVC reward. The resulting architecture, \textit{MVCHead}, is an end-to-end differentiable pipeline that enforces MVC through structural design and a learned geometric reward, rather than relying on explicit 3D supervision. \textbf{(1)} MVCHead comprises a stack of HiSS blocks that progressively refine the Gaussian representation from coarse to fine. These blocks employ the proposed HiBiSS to propagate geometric and appearance cues across a token grid, ensuring local and global consistency when regressing the 3D Gaussian head. \textbf{(2)} The resulting set of 3D Gaussians is processed by a 3DGS rasterizer~\cite{kerbl20233d}. This allows us to render the avatar from arbitrary camera poses. \textbf{(3)} During training, these renders are evaluated by two distinct critics: an adversarial texture discriminator that ensures high-frequency realism and stylistic alignment with the training distribution; and an SE(3) Multi-view Critic that enforces MVC by rewarding pixel-aligned cross-view agreement.

\subsection{Hierarchical State Space (HiSS) Blocks}

We represent the head as a composition of Gaussians that are progressively refined across a hierarchy of $L$ HiSS blocks. Unlike conventional 3DGS~\cite{kerbl20233d}, here Gaussians serve a dual role: they provide a partial, coarse approximation of the 3D head and simultaneously guide the regression of subsequent finer-level Gaussians.\vspace{0.25em}

\noindent {\bf Anchor-based Refinement.} Fine-level Gaussians are parametrized explicitly as offsets from coarser-level anchors~\cite{hyun2024gsgan}. This architectural bias ensures that new primitives lie near established structure, forcing details to refine existing geometry rather than drifting arbitrarily. As synthesis progresses, the Gaussian count grows by an upsampling ratio $r$ per block, enabling progressively detailed synthesis of facial features. Specifically, each subsequent block upsamples its input points~\cite{zhao2021point, guo2021pct, nichol2022point} and attaches new Gaussians to existing ones. The final avatar is rendered jointly in a single splatting pass using the aggregated set of $\sum_{l=0}^{L-1}Nr^{l}$ primitives.\vspace{0.25em}

\noindent {\bf Conditioning and Disentanglement.} The initial HiSS block ($l=0$) takes as input a scaffold of randomly initialized learnable tokens of size $512 \times 3$~\cite{tang2023dreamgaussian}. To increase the representational capacity, these tokens are lifted to a higher-dimensional feature grid via multi-frequency positional encoding, yielding a dense $H \times W$ grid. To ensure identity-consistent synthesis, we apply disentangled appearance conditioning via AdaIN layers~\cite{huang2017arbitrary}. Tokens are modulated by a learned scale and bias predicted from a mapped latent $w \in W$, which empirically helps decouple appearance from geometry throughout the hierarchy. The same conditioning is applied to all HiSS blocks, ensuring appearance coherence while geometry is refined. Notably, following CGSGAN~\cite{barthel2025cgs}, we explicitly omit camera conditioning within these HiSS blocks. By introducing camera poses only during rendering and the SE(3) Multi-view Critic, we prevent the model from collapsing into view-specific 2D heuristics and ensure the MVC signal remains anchored to the 3D geometry.
\vspace{0.5em}

\noindent {\bf Dual-Mixer Architecture.} Within each HiSS block, tokens pass through two complementary mixers: a self-attention that aggregates global semantics and captures long-range dependencies not strongly tied to spatial axes (such as overall facial identity or global cues), and a state space block that enforces local grid-aligned coherence along horizontal and vertical directions via scanning mechanisms described below. The output tokens are then fed to per-attribute MLP heads that directly regress the Gaussian parameters. HiSS blocks operate on a fixed-resolution token grid at all levels, preserving spatial coherence.

\begin{insightbox}
    {1: Axis-aligned Multi-view Drift}
    Multi-view inconsistencies in 3D Gaussian heads are strongly aligned with the image axes. \textit{Yaw} changes mainly cause horizontal shifts: silhouettes move left-right, ear visibility changes, jawlines shift laterally, and fine structures such as hair appear to translate sideways. In contrast, \textit{Pitch} changes mainly produce vertical drift: shading on the nose, cheekbones, eyebrows, and chin shift up-down, and the apparent heights of facial features change. Since most training data normalize heads to be upright and centered, the dominant components of multi-view inconsistency lie along the horizontal and vertical image directions.
\end{insightbox}

\subsection{Hierarchical Bi-directional State Space Scanning (HiBiSS)}
SSMs offer a natural mechanism for imposing architectural constraints along the specific axes where multi-view inconsistencies typically manifest. However, standard unidirectional scans (i.e., left-to-right)~\cite{gu2024mamba} are insufficient for 3D head generation as they lack vertical propagation and introduce causal biases that prevent global context integration. We therefore introduce HiBiSS, which applies four complementary 2D scans: row-wise left-to-right ($\rightarrow$), row-wise right-to-left ($\leftarrow$), column-wise top-to-bottom ($\downarrow$), and column-wise bottom-to-top ($\uparrow$), creating bidirectional recurrent paths that connect any two tokens along both axes. We implement it by adapting SS2D~\cite{liu2024vmamba} to the hierarchical Gaussian prediction setting: tokens are linearly projected, reshaped into an $H \times W$ grid, processed by four symmetric scan trajectories, fused, and re-projected back to the original token space, preserving one-to-one correspondence between spatial positions and token identities.
\vspace{0.25em}

\noindent {\bf Motivation.} Consider a camera with intrinsics $\mathbf{K} = \mathrm{diag}(f_x, f_y, 1)$ and a canonical pose $(R=I, \ t=\mathbf{0})$. A 3D point $X = (X, Y, Z)^\top$ on the head surface projects to pixel coordinates: $\mathbf{u} = (x, y)^\top = \Big(f_x \tfrac{X}{Z},\; f_y \tfrac{Y}{Z}\Big)^\top$. Small yaw and pitch rotations about the vertical and horizontal axes, with angles $\delta\theta_y$ and $\delta\theta_x$ respectively, induce a first-order displacement: $\delta \mathbf{u} \;\approx\; J_x(X)\,\delta\theta_x \;+\; J_y(X)\,\delta\theta_y,$ where $J_x(X) = \tfrac{\partial \mathbf{u}}{\partial \theta_x}$ and $J_y(X) = \tfrac{\partial \mathbf{u}}{\partial \theta_y}$ are the pitch and yaw Jacobians at $X$. For upright, centered heads, where depth $Z$ varies smoothly and the face is approximately centered on the optical axis, we typically observe: $\big|\tfrac{\partial x}{\partial \theta_y}\big| \gg \big|\tfrac{\partial y}{\partial \theta_y}\big|, \big|\tfrac{\partial y}{\partial \theta_x}\big| \gg \big|\tfrac{\partial x}{\partial \theta_x}\big|$, i.e., yaw mainly produces horizontal displacement, while pitch produces vertical displacement. This motivates encoding cross-view corrections with state-space recurrences aligned to rows and columns. 
\vspace{0.25em}

\noindent {\bf HiBiSS Architecture.} Based on this motivation, we introduce HiBiSS to encode cross-view corrections using state space recurrences aligned to the rows and columns. Let $F \in \mathbb{R}^{H \times W \times d}$ denote the 2D token grid, with row index $i \in \{1,\dots,H\}$, column index $j \in \{1,\dots,W\}$, and channel dimension $d$. The horizontal forward scan along row $i$ is defined by the recurrence: 
\vspace{-0.25em}
\[
h^{\rightarrow}_{i,j+1} = A_h\, h^{\rightarrow}_{i,j} + B_h\, F_{i,j}, 
\quad
\tilde{F}^{\text{hor}}_{i,j} = C_h\, h^{\rightarrow}_{i,j} + D_h\, F_{i,j},
\vspace{-0.5em}
\]
where $h^{\rightarrow}_{i,j} \in \mathbb{R}^d$ is the hidden state at position $(i,j)$ and $A_h, B_h, C_h, D_h \in \mathbb{R}^{d \times d}$ are structured state space matrices following the parameterization of \cite{liu2024vmamba}. The vertical forward scan is defined analogously. HiBiSS runs all four directional scans hierarchically and fuses the resulting features into an updated grid $\tilde{F}$. Thus, state-space propagation is explicitly aligned with the directions where $\|\partial \mathbf{u} / \partial \theta\|$ is largest, implementing an anisotropic, pose-aware smoothing that targets the principal axes of inconsistency drift. HiBiSS is applied \emph{before} per-level upsampling and attribute regression. Applying it after upsampling would increase compute and dilute the recurrence over near-duplicate tokens, while applying it during per-attribute prediction would deprive the model of a shared, geometry-aware context. Since the \texttt{Attn+MLP} mixer operates on the same appearance-conditioned features, and passing them through HiBiSS beforehand enables coherent propagation of both appearance and geometric cues, improving multi-view agreement across the full set of Gaussian attributes.

\begin{figure*}[t]
    \centering
    \includegraphics[width=\textwidth]{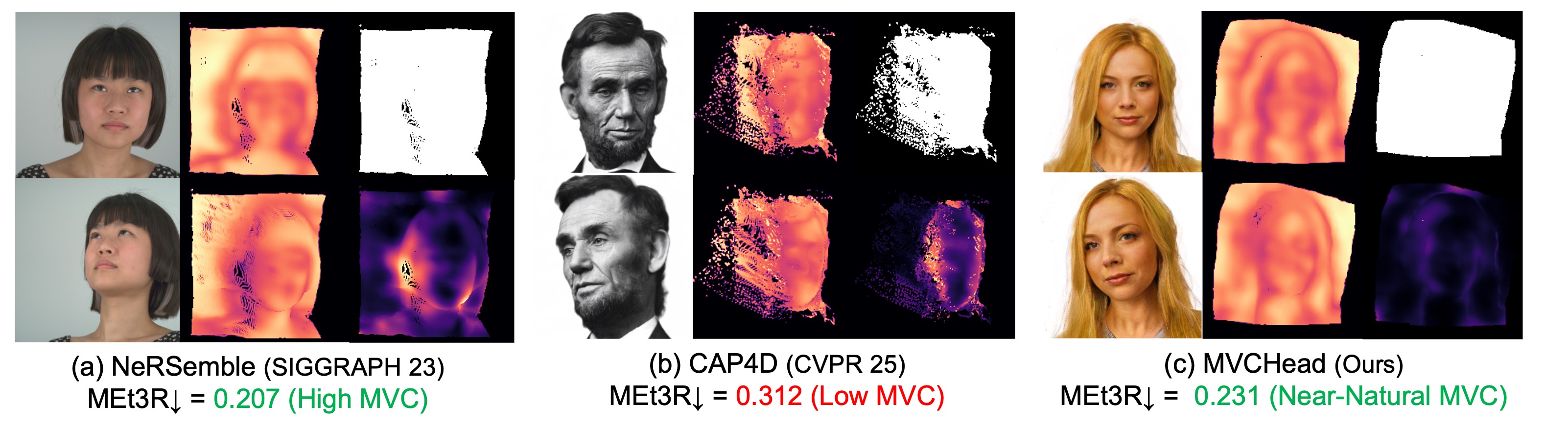}
     \vspace{-2em}
     \captionof{figure}{\textbf{Self-Renders provide strong MVC prior.} We evaluate MVC between view pairs from (a) studio-captured data~\cite{kirschstein2023nersemble}, (b) intermediate view synthesis~\cite{taubner2025cap4d}, and (c) self-renders from 3D. Using MASt3R~\cite{leroy2024grounding} for estimating epipolar-consistent correspondence and FeatUp-DINO~\cite{fu2024featup, caron2021emerging} for measuring feature agreement with a view-invariant encoder, we compute a per-pixel consistency score map over the overlapping region. For each case, we visualize: \textit{Left:} inputs; \textit{Middle:} reprojected views A$\rightarrow$B and B$\rightarrow$A; \textit{Right:} overlap mask and consistency map (dark = consistent, bright = inconsistent). MEt3R~\cite{asim2025met3r} is the spatial average of the error.}
     \vspace{-0.5em}
    \label{fig:4_insight2}
\end{figure*}

\begin{insightbox}
    {2: Consistency by Construction}
    While intermediate view generation may produce inconsistent views, self-renders from any fixed 3D model are, \textit{by construction}, geometrically consistent (see Fig.~\ref{fig:4_insight2}). This allows us to train a critic that distinguishes plausible 3D configurations from inconsistent ones without requiring explicit multi-view data.
\end{insightbox}

\subsection{SE(3) Multi-view Critic}

The Critic is an extrinsic-aware encoder $E_\psi$ that maps a set of images and corresponding camera poses to a scalar consistency score $s=E_{\psi}(\{\hat{I}_k\},\{T_k\}) \in \mathbb{R}$. For a given latent $z$, we render $K$ views $\{\hat{I}_k\}_{k=1}^K$ of the generated avatar under a set of canonicalized camera poses $\{T_k\}_{k=1}^K$. The Critic jointly processes both images and poses to produce a score that is higher when the set is mutually consistent. During training, the model maximizes this score, so that improving multi-view agreement directly improves the objective:
\begin{equation*}
    \begin{aligned}
    \label{eq:1}
    \mathscr{L}_{mvc} = -\mathbb{E}_{z, \{T_k\}}\big[E_\psi\big(\{\mathcal{R}(\mathcal{S}_{\theta}(z),T_k)\}_{k=1}^{K},\; \{T_k\}_{k=1}^{K}\big)\big]
\end{aligned}
\end{equation*}

\noindent {\bf Training Strategy.} To ensure that $E_\psi$ provides a meaningful MVC signal, we train it as a binary set classifier. The positive set $\mathcal{S}^+=\{(\mathcal{R}(\mathcal{S}_\theta(z), T_k), T_k)\}_{k=1}^{K}$ consists of $K$ views rendered from the \textit{same} avatar under different poses $T_k$. The negative set $\mathcal{S}^-=\{(\mathcal{R}(\mathcal{S}_\theta(z_k), T_k), T_k)\}_{k=1}^{K}$ comprises views each rendered from a different latent but sharing the same $T_k$'s. The Critic is optimized with a binary cross-entropy loss on its logits, encouraging it to assign higher scores to positive sets than to negative ones. Although the negative sets exhibit obvious identity variation, the Critic must additionally learn subtle geometric and textural cues of consistency such as silhouette coherence and shading continuity. Once trained, $E_\psi$ serves as a differentiable reward term: the HiSS blocks are updated to maximize $E_\psi(\mathcal{S}^+)$, pushing the model to produce avatars whose self-renders exhibit stronger cross-view consistency.
\vspace{0.25em}

\noindent {\bf Geometric Transform Attention.} The Critic's consistency score should depend only on the relative view arrangement, not absolute camera placement or intrinsics. While standard cross-attention lacks this invariance, Geometric Transform Attention (GTA)~\cite{miyato2023gta} addresses it by embedding SE(3) structure directly into the attention, ensuring equivariance to global rigid transforms and invariance to intrinsics. Architecturally, $E_\psi$ follows a ViT-style design augmented with GTA~\cite{miyato2023gta}. Each image is patchified into tokens. We inject extrinsics by anchoring all poses relative to the first view $\tilde{T}_k = T_k T_1^{-1}$, and align tokens across views by pre-transforming the attention queries and keys with lightweight, block-diagonal linear maps derived from these relative extrinsics. Since GTA aligns tokens using SE(3) relations, i.e., without camera intrinsics, the score $s$ is invariant to intrinsics and cropping, and stable across rig changes, yielding pose-only invariance. Moreover, since the scene and all cameras undergo the same transform, the set of relative transforms is unchanged, $s$ is preserved, providing global-rigid equivariance.
\vspace{0.25em}

\noindent \textbf{Training Objective.} The total loss for MVCHead is a multi-task objective that combines geometric consistency, textural realism, and structural regularization. Given only 2D images, the joint model optimizes the parameters of the Gaussian decoder (HiSS blocks), the SE(3) Multi-view Critic ($E_{\psi}$), and the adversarial texture discriminator ($D_{\phi}$). This joint training pushes the model to produce 3D configurations that are both multi-view consistent and statistically indistinguishable from real images. 

The total loss combines: {\bf (1) SE(3) Multi-view Critic ($\mathscr{L}_{mvc}$):} Encourages cross-view geometric consistency. This constitutes a key departure from prior works such as GS-GAN~\cite{hyun2024gsgan} and CGSGAN~\cite{barthel2025cgs}, which rely primarily on adversarial and conditional losses. {\bf (2) Adversarial texture term ($\mathscr{L}_{adv}$):} A standard camera-conditioned adversarial loss with an R1 gradient penalty. It ensures that the projected textures of generated avatars match the distribution of real training images across $K$ sampled views~\cite{barthel2025cgs}. {\bf (3) Spatial regularization ($\mathscr{L}_{knn}$ and $\mathscr{L}_{ctr}$):} These constrain the Gaussian point cloud~\cite{hyun2024gsgan}. $\mathscr{L}_{knn}$ penalizes excessive spacing between neighboring Gaussians to maintain surface density, while $\mathscr{L}_{ctr}$ penalizes Gaussian centers' drift from their hierarchical anchors to ensure structural stability. The combined loss is defined as follows:
\begin{equation*}
\begin{aligned}
\label{eq:2}
\mathscr{L}_{\text{total}} & =
\underbrace{\lambda_{mvc}\Big(-\mathbb{E}_{z,\{T_k\}}\big[E_\psi(\{\hat{I}_k\},\{T_k\})\big]\Big)}_{\text{SE(3) Multi-view Critic}}\\
&+ \underbrace{\mathbb{E}_z \tfrac{1}{K} \sum\nolimits_{k=1}^K \text{softplus}\big( -D_{\phi} (\mathcal{R}(\mathcal{S}_{\theta}(z),T_k), T_k) \big)}_{\text{Camera-conditioned Adv., K-view AVG. for Texture Consistency}} \\
&+ \underbrace{\lambda_{knn}\mathscr{L}_{knn} + \lambda_{ctr}\mathscr{L}_{ctr}}_{\text{Gaussian Regularizer (Local spacing, Center drift)}}
\end{aligned}
\end{equation*}
\vspace{-0.75em}

\section{Experiments and Results}

\noindent \textbf{Datasets.} For a fair comparison, we train MVCHead under the established experimental protocol on the FFHQ~\cite{karras2019style} and FFHQ-C~\cite{barthel2025cgs} datasets, and benchmark against SOTA generative 3D head models trained on the same datasets.
\vspace{0.25em}

\noindent \textbf{Evaluation Metrics.} Following prior works, we report Fréchet Inception Distance (FID) and FID$_{\text{3D}}$ to measure the perceptual realism of generated avatars. However, these metrics fail to capture MVC. Quantitative evaluation of MVC remains an open challenge in 3D head synthesis, with no universally accepted metrics currently. To address this gap, we adapt scores from two SOTA frameworks: MVGBench~\cite{xie2025mvgbench} and MEt3R~\cite{asim2025met3r}, providing the first comprehensive quantitative assessment of MVC for 3D head avatars.

\vspace{0.25em}

\noindent \textbf{Training.} MVCHead was trained for 10M steps on FFHQ and FFHQ-C using Adam optimizer on 4 NVIDIA H100 GPUs over 3 days. Additional training details, including hyperparameters, are presented in the supplementary.
\vspace{0.25em}

\subsection{Main Results}

We evaluate MVCHead by training independently on FFHQ~\cite{karras2019style} and FFHQ-C~\cite{barthel2025cgs} to ensure a fair comparison. The synthesized avatars demonstrate SOTA visual quality, capturing fine facial features such as wrinkles, hair wisps, and skin blemishes (see Fig.~\ref{fig:1_teasor}).
\vspace{0.25em}

\noindent\textbf{Realism (FID$\downarrow$, FID$_{\text{3D}}\downarrow$).} 
We use FID to assess the perceptual realism of rendered views from generated 3D Gaussian head avatars. FID measures the distributional similarity in the Inception-V3 feature space over 50K renders. Since standard FID evaluates only near-frontal views, we also report FID$_{\text{3D}}$~\cite{barthel2025cgs}, in which camera poses are randomly sampled across a wider range of viewpoints to probe realism under arbitrary viewing angles. The results are summarized in Table~\ref{tab:02_FID} and~\ref{tab:03_FID}. Under this minimal-resource setting, MVCHead achieves SOTA scores, producing visually coherent renders that remain plausible across diverse synthetic identities and viewpoints.
\vspace{0.25em}

\begin{table}[t]
\caption{\textbf{Perceptual Realism.} Comparison of FID scores. $512 \times 512$ resolution was used for the experiments. {$\dag$}Uses super-resolution network. *We report the results from the original paper.\vspace{-0.5em}}
\centering
\resizebox{\columnwidth}{!}{
\begin{tabular}{l|c|cc}
\toprule
\multirow{2}{*}{Method} & \multirow{2}{*}{Venue} & \multicolumn{2}{c}{FID$\downarrow$} \\
 &  & FFHQ~\cite{karras2019style} & FFHQ-C~\cite{barthel2025cgs}\\
\midrule
StyleSDF~\cite{or2022stylesdf} & CVPR 2022 & 11.2$^{\dag}$ & -\\
EpiGRAF~\cite{skorokhodov2022epigraf} & NeurIPS~2022 & 9.92 & -\\
VoxGRAF~\cite{schwarz2022voxgraf} & NeurIPS~2022 & 9.0 & -\\
GMPI~\cite{zhao2022generative} & ECCV~2022 & 8.29 & -\\
StyleNeRF~\cite{gu2021stylenerf} & ICLR 2022 & 7.80$^{\dag}$ & -\\
EG3D~\cite{chan2022efficient} & CVPR~2022 & \sbest{4.70}$^{\dag}$ & -\\
Mimic3D~\cite{chen2023mimic3d} & ICCV~2023 & 5.37 & -\\
GSGAN~\cite{hyun2024gsgan} & NeurIPS~2024 & 5.60 & \tbest{5.17}\\
GSM~\cite{abdal2024gaussian} & CVPR~2024 & 28.19 & -\\
GGHead~\cite{kirschstein2024gghead} & SIGGRAPH Asia~24 & 5.15* & 5.37\\
CGSGAN~\cite{barthel2025cgs} & NeurIPS~2025 & \tbest{4.94} & \sbest{4.53}\\
\textbf{MVCHead} & \textbf{Ours} & \best{\textbf{4.39}} & \best{\textbf{3.94}}\\
\bottomrule
\end{tabular}
}
\label{tab:02_FID}
\end{table}

\begin{table}[t]
\caption{\textbf{Perceptual Realism at extremes.} Comparison of FID$_{\text{3D}}$ scores. $512 \times 512$ resolution was used for the experiments.\vspace{-0.5em}}
\centering
\resizebox{\columnwidth}{!}{
\begin{tabular}{l|c|cc}
\toprule
\multirow{2}{*}{Method} & \multirow{2}{*}{Venue} & \multicolumn{2}{c}{FID$_{\text{3D}}\downarrow$} \\
 &  & FFHQ~\cite{karras2019style} & FFHQ-C~\cite{barthel2025cgs}\\
\midrule
GSGAN~\cite{hyun2024gsgan} & NeurIPS~2024 & 10.50 & \tbest{7.68}\\
GGHead~\cite{kirschstein2024gghead} & SIGGRAPH Asia~24 & \tbest{7.90} & 7.78\\
CGSGAN~\cite{barthel2025cgs} & NeurIPS~2025 & \sbest{4.94} & \sbest{4.53}\\
\textbf{MVCHead} & \textbf{Ours} &  \best{\textbf{4.39}} & \best{\textbf{3.94}}\\
\bottomrule
\end{tabular}
}
\vspace{-1em}
\label{tab:03_FID}
\end{table}

\noindent \textbf{Shape Consistency (CD$\downarrow$, depth$\downarrow$).} We assess shape consistency using Chamfer Distance and depth error. For each generated identity, we construct two independent 3DGS representations, $G_1$ and $G_2$, by optimizing from two disjoint subsets of multi-view renders produced by the same avatar. Each 3DGS is then downsampled to a fixed-budget point cloud of 60K points, yielding $P_1$ and $P_2$, and we compute $e_{\text{cd}}(G_1,G_2) = d_{\text{CD}}(P_1,P_2)$. In addition, we render $K$ depth maps $\pi_i^d(G)$ per 3DGS and measure a masked depth error $e_d$ over the overlapping foreground regions across views. Intuitively, $e_{\text{cd}}$ captures global shape discrepancies, while $e_d$ is sensitive to local errors along silhouettes and fine structures. As reported in Table~\ref{tab:01_MVC}, MVCHead achieves lower CD, indicating improved global shape consistency. The depth error is comparable between the two methods, suggesting that local depth accuracy is similar.
\vspace{0.25em}

\begin{table}[t]
\caption{\textbf{Multi-view Consistency.} Consistency scores of the synthesized 3D Gaussian heads averaged over 100 avatars. \vspace{-0.5em}}
\centering
\resizebox{\columnwidth}{!}{%
        \setlength{\tabcolsep}{1.25pt}
\begin{tabular}{l|cc|ccc|c}
\toprule
\multicolumn{1}{c|}{\multirow{2}{*}{Method}} & \multicolumn{2}{c|}{\textbf{Shape}~\cite{xie2025mvgbench}} & \multicolumn{3}{c|}{\textbf{Texture}~\cite{xie2025mvgbench}} & \textbf{Geometric}~\cite{asim2025met3r}\\
\cmidrule(lr){2-7}
& CD$\downarrow$ & depth$\downarrow$ & cPSNR$\uparrow$ & cSSIM$\uparrow$ & cLPIPS$\downarrow$ & MEt3R$\downarrow$\\
\midrule
CGSGAN~\cite{barthel2025cgs} & 0.6724 & \best{\bf 6.6624} & 21.852 & 0.7434 & 0.0622 & 0.2814\\
\textbf{MVCHead} & \best{\bf 0.6654} & 6.6649 & \best{\bf 22.082} & \best{\bf 0.7636} & \best{\bf 0.0528} & \best{\bf 0.2620} \\
\bottomrule
\end{tabular}
}
\vspace{-1em}
\label{tab:01_MVC}
\end{table}

\noindent \textbf{Texture Consistency (cPSNR$\uparrow$, cSSIM$\uparrow$, cLPIPS$\downarrow$).} To evaluate cross-view texture stability, for each avatar we fit \emph{two} independent 3DGS representations, $G_1$ and $G_2$, from \emph{disjoint} multi-view subsets rendered from the same underlying avatar. We then render each 3DGS into $K$ RGB images $\pi_i(G_1)$ and $\pi_i(G_2)$ under a fixed camera rig and compute MVC metrics between corresponding views: $e_m(G_1, G_2) = \frac{1}{K}\sum_{i=1}^K d_m\big(\pi_i(G_1), \pi_i(G_2)\big)$, where $m \in \{\text{cPSNR}, \text{cSSIM}, \text{cLPIPS}\}$ and $d_m$ denotes the corresponding image-space metric. Since $G_1$ and $G_2$ are reconstructed from non-overlapping view subsets, they coincide only when textures are self-consistent across viewpoints; discrepancies reveal cross-view texture drift. These metrics quantify how well fine texture patterns, such as eyebrows, lip color, skin blemishes, and hair edges, remain stable under pose changes. As reported in Table~\ref{tab:01_MVC}, MVCHead exhibits strong texture consistency.
\vspace{0.25em}

\noindent \textbf{Geometric Consistency (MEt3R$\downarrow$).} To evaluate geometric consistency under larger camera changes, we adopt MEt3R~\cite{asim2025met3r}, which measures MVC directly between image pairs without requiring known camera poses or 3D ground truth. Given a pair of self-renders $(I_1, I_2)$ of a single avatar, we first use MASt3R~\cite{leroy2024grounding} to obtain dense, pose-free stereo reconstructions $X_1, X_2 \in \mathbb{R}^{H\times W \times 3}$ in the coordinate frame of $I_1$. We then extract semantic features with DINO~\cite{caron2021emerging} and upsample them with FeatUp~\cite{fu2024featup} to obtain high-resolution feature maps $F_1$ and $F_2$. Using the MASt3R point maps, these features are unprojected into 3D and reprojected into the frame of $I_1$, yielding aligned feature maps $\hat F_1$ and $\hat F_2$. A masked, pixel-wise cosine similarity between $\hat F_1$ and $\hat F_2$ over the overlapping region defines a directional consistency score $S(I_1, I_2)$. The final MEt3R$(I_1,I_2)$ score is computed as $1 -0.5\cdot\big(S(I_1,I_2) + S(I_2,I_1)\big)$. We adapt this pipeline to head avatars by sampling camera pairs uniformly along yaw and pitch around a canonical rig and computing MEt3R over many random view pairs per identity. As reported in Table~\ref{tab:01_MVC}, MVCHead achieves a lower MEt3R score than SOTA, indicating stronger geometric consistency under large pose changes.

\subsection{Ablation Study}

To verify the effectiveness of each component, we perform an ablation study (see Table~\ref{tab:03_ablation_table}). Removing the adversarial loss $\mathscr{L}_{adv}$ leads to training collapse, confirming its necessity for maintaining image realism. Dropping the MVC loss $\mathscr{L}_{mvc}$ degrades both FID and MEt3R, underscoring the importance of the SE(3) Multi-view Critic for enforcing cross-view consistency. Removing \texttt{SS2D+LN+FFN} (i.e., the state space component) from each HiSS block results in a noticeable decline in MVC, confirming that the SSM contributes meaningfully beyond what attention alone provides. Finally, replacing HiBiSS with a standard unidirectional scan degrades performance, validating that axis-aligned, bidirectional recurrence is critical for reconciling multi-view drift.

\begin{table}[t]
    \centering
    \caption{\textbf{Ablation Study.} Performed on the FFHQ-C~\cite{barthel2025cgs} dataset with $512 \times 512$ resolution to verify the proposed components.~\vspace{-0.5em}}
    \begin{tabular}{clrr}
        \toprule
         & Method & FID$\downarrow$ & MEt3R$\downarrow$~\cite{asim2025met3r} \\
        \midrule
            & Full Model & \textbf{3.94} & \textbf{0.2620} \\
            \midrule
            \multirow{2}{*}{\rotatebox[origin=r]{90}{\parbox[c]{0.4cm}{\centering \small Loss}}}
            & \ \ \  w/o $\mathscr{L}_{adv}$ & \multicolumn{2}{c}{collapse} \\
            & \ \ \ w/o $\mathscr{L}_{mvc}$ & 5.41 & 0.3144 \\
            \cmidrule(lr){1-4}
            \multirow{2}{*}{\rotatebox[origin=r]{90}{\parbox[c]{0.5cm}{\centering \small Model Arch.}}}
            & \ \ \ w/o HiSS block  & 5.28 & 0.2948 \\
            & \ \ \ w/o HiBiSS    & 4.78 & 0.2873 \\
        \bottomrule
    \end{tabular}
    \label{tab:03_ablation_table}
\end{table}

\subsection{Extensions}

\noindent \textbf{FaceGS-10K Dataset.} 
To demonstrate a direct application of unconditional 3D head generation at scale, we construct \textit{FaceGS-10K}—to our knowledge, the first large-scale dataset of ready-to-use 3D Gaussian head assets that is independent of any parametric 3D head model. Each asset contains $240$K anisotropic Gaussians, along with 24 renderings over the frontal hemisphere at a resolution of $512\times512$. We generate the dataset by sampling diverse latent codes from the trained MVCHead model, retaining only identities that meet both a cross-view consistency threshold and a frontal realism filter. In contrast to purely 2D datasets~\cite{karras2019style}, multi-view image collections without underlying 3D representations~\cite{kirschstein2023nersemble}, or FLAME-registered meshes~\cite{yang2020facescape, li2017learning}, FaceGS-10K stores raw Gaussian attributes that can be directly rendered using off-the-shelf 3DGS rasterizers~\cite{kerbl20233d}. \textit{FaceGS-10K} can support a range of downstream applications, including providing 3D supervision for reconstruction methods, and enabling privacy-preserving synthetic identity generation for AR/VR and content creation. More details in the supplementary material.
\vspace{0.25em}

\noindent \textbf{Conditional Generation.}  We adapt MVCHead for personalized avatar creation from a single image using optimization-based inversion. Given an input face image, we recover a latent code and minimize an ArcFace-based identity preservation loss~\cite{deng2019arcface}. Because the architecture remains unchanged, multi-view consistency (MVC) is naturally preserved in the personalized results. We emphasize that conditional generation is not the primary focus of this work; rather, it demonstrates that MVCHead is sufficiently structured to support inversion while preserving MVC. More details in the supplementary material.

\section{Conclusion and Future Work}

We present MVCHead, the \textit{first} state space model for 3D Gaussian heads designed to address multi-view consistency (MVC) in the minimal-resource setting. MVCHead generates high-fidelity, multi-view consistent 3D head avatars in a single forward pass, achieving SOTA performance on five of six metrics. At its core are the HiSS block, which aligns SSM recurrence with the principal axes of drift via HiBiSS scanning, and the SE(3) Multi-view Critic, which enhances MVC by \emph{design} without studio data or intermediate view synthesis. To our knowledge, this is the first comprehensive analysis of multi-view consistency in 3D head avatars.
\vspace{0.25em}

\noindent\textbf{Limitations.} Despite its strengths, MVCHead has several limitations. First, it is only trained on front and side views, and cannot generate full 360° avatars; future work could add back-of-head coverage. Second, its geometric priors are learned entirely from 2D supervision. More explicit structural constraints (e.g., bilateral symmetry) could further reduce the search space. Additionally, harder negatives for the critic, e.g., geometrically perturbed views of the same identity, can further strengthen the consistency signal.
\vspace{0.25em}

\noindent \textbf{Acknowledgments}. The computational resources were supported by PSC Bridges-2 through the Advanced Cyberinfrastructure Coordination Ecosystem: Services and Support (ACCESS) program allocation CIS250961. The authors thank Francisco Vicente Carrasco, Saswat Subhajyoti Mallick, Jianjin Xu, and José Pedro Gomes for their suggestions and feedback that improved the work.